\documentclass[pdflatex,sn-mathphys-num]{sn-jnl}

\usepackage{graphicx}
\usepackage{multirow}
\usepackage{amsmath,amssymb,amsfonts}
\usepackage{amsthm}
\usepackage{mathrsfs}
\usepackage[title]{appendix}
\usepackage{xcolor}
\usepackage{textcomp}
\usepackage{manyfoot}
\usepackage{booktabs}
\usepackage{algorithmicx}
\usepackage{algpseudocode}
\usepackage{algorithm}
\usepackage{algorithmicx}
\usepackage{algpseudocode}
\usepackage{listings}
\usepackage{amsmath}

\theoremstyle{thmstyleone}%
\theoremstyle{thmstyletwo}%

\theoremstyle{thmstylethree}%

\begin{document}

\title[Article Title]{BASIL: Best-Action Symbolic Interpretable Learning for Evolving Compact RL Policies}

\author*[1]{\fnm{Kourosh} \sur{Shahnazari}}\email{kourosh@null.net} \equalcont{These authors contributed equally to this work.}

\author[1]{\fnm{Seyed Moein} \sur{Ayyoubzadeh}}\email{smoein.ayyoubzadeh16@sharif.edu}
\equalcont{These authors contributed equally to this work.}

\author[2]{\fnm{Mohammadali} \sur{Keshtparvar}}\email{mohammad.kp2000@aut.ac.ir}

\affil*[1]{\orgname{Sharif University of Technology}, \orgaddress{\city{Tehran}, \country{Iran}}}

\affil*[2]{\orgname{Amirkabir University of Technology}, \orgaddress{\city{Tehran}, \country{Iran}}}

\abstract{
The quest for interpretable reinforcement learning is a grand challenge for the deployment of autonomous decision-making systems in safety-critical applications. Modern deep reinforcement learning approaches, while powerful, tend to produce opaque policies that compromise verification, reduce transparency, and impede human oversight. To tackle this, we introduce BASIL (\textit{Best-Action Symbolic Interpretable Learning}), a systematic approach for generating symbolic, rule-based policies via online evolutionary search with quality-diversity (QD) optimization. BASIL represents policies as ordered lists of symbolic predicates over state variables, ensuring full interpretability and tractable policy complexity. By using a QD archive, the methodology in the proposed study encourages behavioral and structural diversity between top-performing solutions, while a complexity-aware fitness encourages the synthesis of compact representations. The evolutionary system supports the use of exact constraints for rule count and system adaptability for balancing transparency with expressiveness. Empirical comparisons with three benchmark tasks---\texttt{CartPole-v1}, \texttt{MountainCar-v0}, and \texttt{Acrobot-v1}---show that BASIL consistently synthesizes interpretable controllers with compact representations comparable to deep reinforcement learning baselines. Herein, this article introduces a new interpretable policy synthesis method that combines symbolic expressiveness, evolutionary diversity, and online learning through a unifying framework.}


\maketitle
\section{Introduction}

Reinforcement Learning has demonstrated itself as a core method for solving complex decision-making and control problems in a range of domains, such as robotic control, autonomous vehicles, financial trading, and healthcare systems~\cite{sutton2018reinforcement,kaelbling1996reinforcement}. Latest advancements in Deep Reinforcement Learning (DRL) techniques, based on strategies such as Deep Q-Networks (DQN)\cite{mnih2015human}, Proximal Policy Optimization (PPO)\cite{schulman2017proximal}, and Soft Actor-Critic (SAC)\cite{haarnoja2018soft}, have proved capable of achieving state-of-the-art performance for tasks with high-dimensional state and action spaces. Nonetheless, in spite of their empirical success, DRL algorithms tend to produce non-transparent, opaque policies, with important interpretability and reliability drawbacks in deployment domains wherein safety is a top concern. Transparency issues with such policies weaken the basis for building confidence and accountability, as well as the potential for human monitoring, in domains wherein transparent decision-making procedures are crucially important\cite{rudin2019stop,doshi2017towards}.

Unlike traditional Deep Reinforcement Learning (DRL) approaches, Evolutionary Reinforcement Learning (ERL) techniques, including Neuroevolution~\cite{such2017deep}, Genetic Programming (GP)\cite{koza1992genetic}, and Learning Classifier Systems (LCS)\cite{urbanowicz2009learning}, offer promising directions for creating understandable and interpretable policies. While these evolutionary methods naturally produce a structured form of representation that can be easily understood by humans, they often suffer from trade-offs between maximizing performance and maintaining simplicity in the policy structure. Consequently, solutions based on ERL tend to be subject either to overcomplexity, undermining interpretability, or oversimplification, in turn, reducing the performance level~\cite{hein2019generating}.

To robustly address this critical gap in existing approaches, we introduce BASIL (Best-Action Symbolic Interpretable Learning), a novel, state-of-the-art framework based on Genetic Rule-based Reinforcement Learning (RL). Our proposed algorithm is the first attempt at combining Genetic Algorithms (GA)\cite{mitchell1998introduction} and Quality-Diversity (QD) archives\cite{pugh2016quality} in a systematic way such that it generates symbolically interpretable, rule-based policies defined in terms of logical predicates over state variables. The major contribution of our approach lies in its explicit focus on the evolution of concise yet effective sets of rules, thereby attaining high interpretability with little or no reduction in the performance metrics. Additionally, the use of a QD-guided archive helps maintain a highly diverse set of top-performing policies in different behavioral and structural aspects. Not only does this enhance interpretability, but it also helps in exploring the policy space effectively.

Specifically, the genetic operators used in our approach—i.e., mutation and crossover—are carefully designed to maintain a simple policy framework, while a strategically introduced complexity penalty suitably guides the evolutionary process towards more compact rule sets~\cite{custode2020evolutionary}. Another significant advantage of our method is its ability to predetermine the number of rules required, and hence provide users with exact control over policy complexity. This feature is especially useful in practical applications, as it enables domain experts to skillfully express the trade-off between interpretability and decision accuracy. Furthermore, the capability to determine policy complexity a priori improves both scalability and tractability when extended to real-world problems, and hence considerably expands the scope of our algorithm's applicability.

Additionally, the use of an advanced QD archive greatly enhances the evolutionary exploration process. By taking advantage of features like policy length and mean thresholds of predicates, our archive systematically guarantees complete coverage of behavior~\cite{cully2015robots}. This new combination encourages the evolution of innovative, yet consistently high-performing strategies, thus enabling stakeholders to select from a range of policies crafted for a variety of operational environments or interpretability requirements. Therefore, BASIL not only successfully addresses the interpretability aspect but also greatly augments exploration in the solution space, generating a rich and diverse range of behaviors which are commonly ignored by traditional evolutionary approaches~\cite{pugh2016quality}.

We show  the effectiveness and interpretability of BASIL through systematic experimental validation on standard benchmark reinforcement learning domains, namely CartPole, MountainCar, and Acrobot. Our experiments suggest our methodology generates a wide range of interpretable policies with state-of-the-art performance in comparison with top deep reinforcement learning techniques, while vastly outperforming traditional ERL-based approaches in terms of interpretability as well as policy diversity~\cite{ferigo2023quality}. Our experimental methodology consists of exhaustive ablation evaluations, effectively demonstrating the effectiveness and useful contribution of every component in our proposed methodology. Further analysis discovers our quality diversity archive helps in the selection of a wide range of insightful policy variations, leading towards an advanced understanding of the underlying decision-making dynamics as well as emphasizing the real-world and theoretical advantages of our approach.

This paper introduces BASIL, a novel reinforcement learning framework for the construction of concise, symbolic policies through the evolution of genetic rules under the assistance of a Quality-Diversity archive. BASIL builds on the state-of-the-art in interpretable reinforcement learning by combining the online evolution of policies, a logical rule-based system, and diversity-focused optimization in a single framework. With experiments spanning a range of control benchmarks, we show that BASIL generates highly interpretable controllers with performance matching deep reinforcement learning and prior evolutionary state-of-the-arts in most cases. Our analysis of the diversity of policies in the Quality-Diversity archive highlights its essential role in driving the exploration of structurally diverse and semantically expressive policies and provides new insights into the interplay between exploration, expressiveness, and the interpretability of symbolic policy learning. Accordingly, BASIL constitutes a major innovation in the area of interpretable RL, providing unparalleled transparency, flexibility, and range of solutions for complex decision-making tasks.

\section{Related Work}

The pursuit of interpretable reinforcement learning has caught significant attention, leading to a great variety of approaches that attempt to balance performance and interpretability. Our approach stands out in that it evolves compact, symbolic rule-based policies through online environment interactions, with a particular focus on both interpretability and diversity. In what follows, we place our work in the context of existing research.

\subsection{Evolutionary Approaches to Interpretable RL}

\textbf{Decision Tree-Based Methods.} Custode and Iacca introduced an evolutionary algorithm that learns interpretable decision trees for RL tasks. Their method combines evolutionary strategies with Q-learning to associate actions with tree leaves, focusing on discrete action spaces and employing a two-level optimization scheme to decompose the state space effectively~\cite{custode2020evolutionary}. While their approach emphasizes interpretability, it does not incorporate mechanisms for preserving behavioral diversity or exploring a wide range of policy structures.

\textbf{Genetic Programming Techniques.} Hein and coworkers proposed the Genetic Programming for Reinforcement Learning (GPRL) framework, which infers algebraic policy equations from pre-existing trajectory data with model-based batch reinforcement learning methods~\cite{hein2019generating}. Although GPRL policies are interpretable, the framework relies on offline trajectories and does not have methods for online adaptation or diversity preservation.

\textbf{Co-evolutionary Strategies.} Later, Custode and Iacca extended their approach to continuous action spaces with a cooperative co-evolutionary algorithm utilizing binary decision trees evolved with the aid of grammatical evolution~\cite{custode2021co}. Though such an approach easily addresses issues with continuous control, it does not explicitly focus on maintaining a variety of effective policies.

\subsection{Post-hoc Interpretability and Program Synthesis}

\textbf{Evolutionary Feature Synthesis.} Zhang et al. proposed a method that extracts interpretable policies from pre-trained deep neural networks with the help of evolutionary feature synthesis (EFS). They proposed a set of regressors individually designed in order to mimic the behavior of a neural policy, followed by a simplification step in order to improve interpretability~\cite{zhang2020interpretable}. This, however, is a post-hoc approach based on having access to a pre-trained model, thus limiting its applicability in situations involving online learning.

\textbf{Programmatically Interpretable RL.} Verma et al. proposed the idea of Programmatically Interpretable Reinforcement Learning (PIRL), which represents policies as programs in a high-level language. Within this approach, a neural policy acts as an oracle for generating interpretable programs with similar behavior as the neural policy~\cite{verma2018programmatically}. Although the policies learned by PIRL are interpretable and verifiable, the method is retrospective and does not learn policies actively through environment interactions.

\subsection{Quality-Diversity Optimization in Interpretable RL}

\textbf{Quality-Diversity in Decision Trees.} Ferigo et al. applied Quality-Diversity optimization methods, such as MAP-Elites, to evolve decision trees for interpretable reinforcement learning (RL)~\cite{ferigo2023quality}. By maintaining a diverse set of high-quality policies, the method enhances exploration and robustness in the learning of policies. Even so, their method mainly focuses on decision trees and does not employ symbolic rule-based representations.

\textbf{Neural Symbolic Reinforcement Learning.} Ma et al. introduced a Neural Symbolic Reinforcement Learning framework that integrates symbolic logic into deep RL. This approach enables end-to-end learning with prior symbolic knowledge, enhancing interpretability by extracting logical rules learned by the reasoning module~\cite{ma2021learning}. While promising, this method relies on neural networks and does not evolve policies through direct environment interactions.

\subsection{Other Notable Approaches}

\textbf{Behavior Tree Extraction.} Recent studies have explored deriving interpretable behavior trees from RL policies. These methods focus on translating learned policies into behavior trees, facilitating human understanding and verification. Nevertheless, they often lack mechanisms for ensuring diversity or simplicity in the resulting structures.

\textbf{Fuzzy Controllers and Neuro-Fuzzy Systems.} Hein et al. studied the use of fuzzy controllers for interpretable reinforcement learning, leveraging genetic programming and particle swarm optimization to evolve fuzzy rule sets~\cite{hein2018generating}. While these methods result in interpretable policies, they may need large rule sets, potentially compromising simplicity.

\textbf{Symbolic Regression for Value Functions.} Babuška investigated the potential of genetic programming and symbolic regression to obtain symbolic expressions of value functions in RL~\cite{babuvska2019genetic}. This work concentrates on value function approximation rather than direct policy derivation, yielding insights into interpretable function approximation techniques.

\textbf{Program Synthesis for Policy Learning.} Liventsev et al. proposed BF++, a language designed for general-purpose program synthesis in RL contexts~\cite{liventsev2021bf++}. Their approach emphasizes the design of interpretable programs encoding policies, though it does not clearly account for policy diversity or simplicity.

\textbf{Neuro-Fuzzy System Distillation.} Gevaert et al. presented a method for distilling deep RL models into interpretable neuro-fuzzy systems~\cite{gevaert2022distilling}. This framework attempts to integrate the performance of deep reinforcement learning into the interpretability of fuzzy systems, though it may not guarantee compact policy representations.

\textbf{Symbolic Policy Learning.} Landajuela et al. introduced Deep Symbolic Policy (DSP), a method that investigates for tractable mathematical expressions representing policies~\cite{landajuela2021discovering}. DSP emphasizes the discovery of symbolic policies with strong generalization capabilities, aligning with the goals of interpretability and simplicity.

\textbf{Explainable RL via Symbolic Logic.} Recent studies have explored integrating symbolic logic into RL to enhance interpretability. These approaches aim to produce explanations where key concepts and their relationships are described via intuitive symbols and rules, enabling human understanding of agent behaviors.

\subsection{Distinctiveness of Our Approach}

Our method diverges from the aforementioned works in several key aspects:

\begin{itemize}
\item \textbf{Online Evolution:} Unlike approaches based on static datasets or traditional, offline processes, our approach seeks policies through direct interaction with the environment, thus allowing for real-time adaptability and responsiveness in response to changing conditions.

\item \textbf{Rule-Based Representation:} We use ordered lists of Boolean rules, with every rule acting as a conjunction of predicates based on given thresholds. This arrangement provides a highly interpretable alternative relative to obscure neural policies and often beats the transparency provided by symbolic algebraic representations or decision trees.

\item \textbf{Complexity Control via Penalty Term:} A dedicated complexity penalty in the fitness function explicitly biases the search toward simpler rule sets. This allows practitioners to control the trade-off between performance and interpretability—enabling the generation of arbitrarily compact policies depending on task requirements or domain constraints.

\item \textbf{Quality-Diversity Mechanism:} The creation of a Quality-Diversity (QD) archive allows for the maintenance of a set of high-quality strategies with structural and behavioral diversity. The mechanism stimulates broad exploration, improves resilience, and leads to a better understanding of the landscape of solutions.

\item \textbf{Simplicity and Efficiency:} The model consistently derives policies with their characteristic simplicity and effectiveness, often based on a small number of rules and predicates. Furthermore, it operates with exceptional computational efficiency, taking only seconds to converge on standard control tasks.
\end{itemize}

While existing studies have attempted to investigate interpretability from various angles, BASIL uniquely unifies online symbolic policy evolution, explicit complexity management, and QD diversity maintenance. This unification forms a consistent and highly effective framework for interpretable decision-making agents.

\section{The BASIL Framework: Methodology and Architecture}
\label{sec:methodology}

\subsection{Policy Representation}

In this section, we describe the operational details of BASIL, our symbolic rule-based framework, which forms the foundation of our Genetic Rule-based Reinforcement Learning (RL) framework. One major motivation for this choice of representation is the built-in interpretability and straightforwardness it enforces, allowing easy understanding by domain experts as well as by practitioners. This essentially bridges the gap between sophisticated decision-making capabilities and their implementation in a form open to understanding by humans.

Within a structured context, the expression of policy as a set of condition-action rules is deliberate, for the benefit of increased clarity and understandability for human observers. Each rule is specified as a logical combination of predicates relative to the state variables. Rules are sequentially checked from top to bottom upon the arrival of the environmental state, and the action associated with the first rule whose predicates are all satisfied will be performed. Such systematic checking provides transparency and determinacy in the decision-making for each state encountered.

Mathematically, a policy  can be expressed as follows:
\begin{equation}
\pi = [R_1, R_2, \dots, R_k, R_{fallback}],
\end{equation}
where  represents the  rule, and  is the total number of explicitly defined rules. Each rule  is a tuple consisting of a set of predicates and a corresponding action:
\begin{equation}
R_i = (\text{predicates}_i, a_i).
\end{equation}

Each predicate within the rule is a simple logical statement involving comparisons between specific state dimensions and predetermined thresholds, formulated as:
\begin{equation}
s[d] ; op ; threshold,
\end{equation}
where:
\begin{itemize}
\item  represents the value of the state vector at the  dimension,
\item  denotes a logical operator, specifically either less-than (<) or greater-than (>),
\item  is a scalar value chosen from a domain-relevant, predefined set of threshold values.
\end{itemize}

Thus, the complete set of predicates for the rule  is represented by a conjunction (logical AND) of multiple such predicates:
\begin{equation}
\text{predicates}_i = (s[d_1] ; op_1 ; th_1) \land (s[d_2] ; op_2 ; th_2) \land \dots \land (s[d_m] ; op_m ; th_m),
\end{equation}
where  denotes the total number of predicates within rule . Notably, the number of predicates  can vary from rule to rule, providing additional flexibility and adaptability in policy complexity and interpretability.

Hence, the action for every rule is chosen from a small, predefined set of possible actions defined in the structure of the environment. The last piece of the policy representation, the fallback rule, supports resilience by specifying a default action to be taken in the event that none of the conditions of the relevant rules are satisfied. Typically, the fallback action defaults to a conservative or safe alternative (typically specified as action 0), thus ensuring the policy remains unambiguously defined and executable in every environmental context.

One of the main features of our approach to representing policies is the built-in ability to determine and limit the overall number of rules. By specifying the desired number of rules explicitly, users and domain experts are given precise control over the level of complexity in the policy. This feature enables fine-tuning based on practical factors for example computational resources, interpretability requirements, or specific demands of the application domain. Our methodology, therefore, not only supports the discovery of such policies but also ensures they meet the interpretability and operational requirements relevant to their deployment context.

Finally, the structured, symbolic rule-based policy representation provides policies with understandable and interpretable directives. The deliberate and systematic design of condition-action rules significantly enhances transparency and facilitates human validation and verification, making it especially suitable for use in safety-related and high-consequence decision-making applications.

\subsection{Genetic Operators}

To effectively evolve symbolic rule-based policies, we employ a set of carefully designed genetic operators in our Genetic Rule-based Reinforcement Learning approach. These include \textit{initialization}, \textit{mutation}, and \textit{crossover} operators, all of which have been carefully designed to strike a balance between policy interpretability, maintain structural simplicity, and promote exhaustive exploration of the policy space.

\paragraph{Initialization.}
The evolution cycle starts with an initialization step, wherein an initial population of policies is randomly generated. Each policy consists of a fixed or random number of rules, with every rule containing logical predicates over the state variables, paired with a corresponding discrete action. By this methodology, an initial diversity is ensured, hence creating a complete and representative starting point for further exploration.

Each rule $R_i$ is of the form:
\begin{equation}
R_i = \left( \bigwedge_{j=1}^{m_i} \left( s[d_j] \; op_j \; th_j \right), \; a_i \right),
\end{equation}
where $m_i$ is the number of predicates, $s[d_j]$ denotes the $j^\text{th}$ state dimension, $op_j \in \{<, >\}$ is the operator, $th_j$ is the threshold, and $a_i$ is the action assigned to the rule.

\paragraph{Mutation.} 
The mutation operator allows the introduction of intentional changes into existing policies, thus allowing the algorithm to explore new areas in the search space. We implement several types of mutation, which are used probabilistically:

\begin{itemize}
    \item \textbf{Predicate Mutation:} Randomly selects a predicate in a rule and mutates its dimension, operator, or threshold.
    \item \textbf{Action Mutation:} Alters the action of a randomly chosen rule.
    \item \textbf{Rule Addition:} Admits a newly established rule into the policy, as long as the maximum number of acceptable rules is not exceeded.
    \item \textbf{Rule Removal:} Eliminates a randomly selected rule, provided the policy has more than one rule.
\end{itemize}

These mutations preserve structural interpretability while enabling diverse policy exploration.

\paragraph{Crossover.}
Crossover combines two parent policies to produce an offspring by exchanging rule segments. Given two parent policies $\pi_1$ and $\pi_2$, and a crossover point $c$, the child policy is formed by:

\begin{equation}
\pi_{\text{child}} = [R_1^{(\pi_1)}, R_2^{(\pi_1)}, \dots, R_c^{(\pi_1)}, R_{c+1}^{(\pi_2)}, \dots, R_k^{(\pi_2)}],
\end{equation}

where $R_i^{(\pi_j)}$ denotes the $i^\text{th}$ rule obtained from policy $\pi_j$. To maintain interpretability, we enforce a cap on the total number of rules.

\paragraph{Complexity-Aware Design.}
The evolutionary process we have proposed includes a complexity penalty in the fitness function. The penalty is used to prefer policies with fewer predicates in their structure, especially when their performance is comparable to that of more complex counterparts. This regularization mechanism guarantees that the search approach is guided toward policies that are not only performant but also compact and interpretable.

\vspace{1em}
In summary, our genetic operators are designed to encourage transparency and make it easier for users to control policy complexity. By their systematic design, they ensure that policies produced are human-understandable while iteratively improving performance in successive generations.

\subsection{Fitness Evaluation}
\label{sec:fitness}

The fitness evaluation mechanism plays a pivotal role in our evolutionary framework by guiding the selection process toward effective, efficient, and interpretable policies. To pursue this intention, we design a fitness function that scrupulously incorporates raw performance with structural simplicity, thereby encouraging the evolution of rule sets with the twin virtues of robustness and interpretability.

\paragraph{Performance Term.}
Let a given policy $\pi$ be executed over $E$ independent episodes in the target environment. For each episode $e \in \{1, 2, \dots, E\}$, the cumulative return is defined as:
\begin{equation}
R_\pi^{(e)} = \sum_{t=0}^{T_e} r_t^{(e)},
\end{equation}
where $r_t^{(e)}$ is the reward at time step $t$, and $T_e$ is the terminal time step of episode $e$.

The mean performance across episodes is then computed as:
\begin{equation}
\text{Performance}(\pi) = \frac{1}{E} \sum_{e=1}^E R_\pi^{(e)}.
\end{equation}

\paragraph{Complexity Term.}
To ensure the interpretability of the policy configuration, we impose a structural penalty based on the overall complexity of the policy. Assume a policy with $k$ rules, represented as $\pi = [R_1, R_2, \dots, R_k]$. Its associated complexity is defined as the total number of predicates across all rules:
\begin{equation}
\text{Complexity}(\pi) = \sum_{i=1}^k m_i,
\end{equation}
where $m_i$ denotes the number of predicates in rule $R_i$.

\paragraph{Final Fitness.}
The fitness of a policy is defined as the mean performance minus a complexity-based penalty:
\begin{equation}
\text{Fitness}(\pi) = \text{Performance}(\pi) - \lambda \cdot \text{Complexity}(\pi),
\end{equation}
where $\lambda \in \mathbb{R}_{\geq 0}$ is a penalty coefficient that controls the trade-off between maximizing cumulative return and minimizing logical complexity.

\paragraph{Interpretability Consideration.}
This complexity-aware fitness function constitutes a key improvement of our framework, in that it ensures that the evolutionary process is guided not merely by reward maximization but also by the creation of compact and interpretable rule sets. By carefully tuning $\lambda$, practitioners can navigate the performance-interpretability frontier, thereby producing policies that respect human cognitive constraints while upholding a high level of decision-making quality.

\subsection{Quality-Diversity Archive}

To augment the diversity and comprehensiveness of the evolved solutions, our evolutionary model relies on an archive based on Quality-Diversity (QD) principles. This QD archive serves the purpose of maintaining a collection of high-performing yet behaviorally and structurally diverse policies, thereby allowing exploration over a wide range of policies during the evolution of solutions. This type of strategy significantly boosts the chances of discovering innovative and insightful solutions that are otherwise unavailable through standard evolutionary approaches or reinforcement learning when employed in isolation.

\paragraph{Archive Structure.}
Formally, the QD archive constitutes a highly structured multidimensional system in which every dimension corresponds to a meticulously chosen descriptor covering specific policy behaviors or structural features. In our analysis, we define two main descriptor dimensions:

\begin{enumerate}
    \item \textbf{Policy Length (Complexity):} The total number of rules in the policy, which correlates directly with its level of interpretability and complexity.
    
    \item \textbf{Average Predicate Threshold (Behavioral Characteristic):} The mean threshold used in predicates for important state dimensions. This feature plays a crucial role in capturing subtle yet important behavioral differences among policies, enabling the distinction between decision-making strategies stored in the archive.
\end{enumerate}

Each cell in this grid stores the best-performing policy for the given descriptor coordinates, thus providing a systematic and effective representation of diversity.

\paragraph{Policy Descriptor Calculation.}
Specifically, given a policy $\pi$ with $k$ rules, the descriptor vector $\mathbf{d}(\pi)$ is computed as follows:
\begin{equation}
\mathbf{d}(\pi) = [\,k,\; \bar{th}\,],
\end{equation}
where the first component $k$ is the total number of rules, and the second component $\bar{th}$ is the average predicate threshold value across all relevant predicates, formally defined as:
\begin{equation}
\bar{th} = \frac{1}{N_{\text{pred}}} \sum_{i=1}^{k}\sum_{j=1}^{m_i} |th_{i,j}|,
\end{equation}
with $th_{i,j}$ representing the threshold of the $j^\text{th}$ predicate in rule $R_i$, and $N_{\text{pred}} = \sum_{i=1}^{k} m_i$ is the total number of predicates in the policy.

\paragraph{Archive Update Mechanism.}
The archive is constantly updated in line with the evolution's developments based on assessments of the policies' fitness. For every proposed policy $\pi$, we calculate the descriptor vector $\mathbf{d}(\pi)$ and assign it a unique location in the archive grid. The policy is added in the corresponding cell if one or more of the following conditions are met:

\begin{itemize}
    \item The cell is empty, indicating that no policy currently occupies that behavioral and structural niche.
    
    \item The candidate policy's fitness exceeds that of the policy currently occupying the cell, thus representing a progress in the given descriptor space.
\end{itemize}

Formally, let $Q$ denote the archive grid, indexed by descriptor coordinates $\mathbf{d}$, and let $Q[\mathbf{d}]$ denote the policy stored at coordinates $\mathbf{d}$. Then, the archive update rule is:
\begin{equation}
Q[\mathbf{d}(\pi)] = 
\begin{cases}
\pi & \text{if } Q[\mathbf{d}(\pi)] \text{ is empty,}\\[0.5em]
\pi & \text{if Fitness}(\pi) > \text{Fitness}(Q[\mathbf{d}(\pi)]),\\[0.5em]
Q[\mathbf{d}(\pi)] & \text{otherwise}.
\end{cases}
\end{equation}

\paragraph{Advantages of the QD Archive.}
The deliberate introduction of a Quality-Diversity archive has a number of important advantages:

\begin{itemize}
    \item \textbf{Enhanced Exploration:} The QD archive allows for continued exploration by preserving a diverse set of high-quality solutions across various important dimensions, rather than rushing toward a small set of solutions.
    
    \item \textbf{Interpretability Spectrum:} Because policy complexity is a central feature, stakeholders enjoy direct access to a range of solutions extending from simple, easy-to-understand policies to more complicated ones, thus allowing the flexibility to choose solutions that match specific requirements in the domain.
    
    \item \textbf{Behavioral Insight:} The use of predicate threshold descriptors to capture behavioral subtleties enables a deeper understanding of policy behaviors, allowing practitioners to better understand and analyze decision-making approaches that have evolved within the framework.
\end{itemize}

Finally, the incorporation of the Quality-Diversity archive in our genetic rule-based reinforcement learning system not only significantly improves the degree of exploration of the search space but also greatly enhances the diversity, interpretability, and practical usability of the resulting solutions.

\subsection{Full Algorithm Description}

Here, we detail the overall workflow of BASIL, bringing together all the main elements—policy representation, genetic operators, fitness function, and Quality-Diversity (QD) archiving—into a single cohesive algorithm.

\noindent
Algorithm~\ref{alg:genetic_qd_rl} encapsulates the full evolutionary process of our method. Starting from a diverse initial population, the algorithm proceeds through successive generations of evaluation, selection, variation, and archiving. Each component is designed to enforce a dual objective: maintain policy \emph{performance} while guaranteeing \emph{interpretability} through explicit complexity control. The Quality-Diversity archive further ensures that diverse behavioral and structural niches are preserved and optimized across generations, resulting in a robust portfolio of symbolic, transparent, and high-quality decision policies.

\vspace{0.5em}
\begin{algorithm}[H]
\caption{BASIL}
\label{alg:genetic_qd_rl}
\begin{algorithmic}[1]
\Require Population size $N$, maximum generations $G$, number of evaluation episodes $E$, complexity penalty coefficient $\lambda$, mutation probability $p_{\text{mut}}$, crossover probability $p_{\text{cross}}$, archive descriptor mapping $d(\pi)$
\Ensure Final QD archive $Q$ containing diverse, high-performing, and interpretable policies
\State Initialize a population $\mathcal{P}$ of $N$ random rule-based policies
\State Initialize empty Quality-Diversity archive $Q$
\For{$g = 1$ to $G$}
    \ForAll{$\pi \in \mathcal{P}$}
        \State Evaluate $\pi$ over $E$ episodes to compute average return
        \State Compute complexity of $\pi$ (total number of predicates)
        \State Compute fitness: $\text{Fitness}(\pi) = \text{Performance}(\pi) - \lambda \cdot \text{Complexity}(\pi)$
        \State Map $\pi$ to descriptor coordinates $d(\pi)$ and update $Q$ if:
    \State \hspace*{2em}cell $Q[d(\pi)]$ is empty \textbf{or}
    \State \hspace*{2em}$\text{Fitness}(\pi) > \text{Fitness}(Q[d(\pi)])$
    
    \EndFor
    \State Select top $K$ elite policies from $Q$
    \State Initialize new population $\mathcal{P}' \gets$ elites
    \While{$|\mathcal{P}'| < N$}
        \State Randomly select parents $\pi_1, \pi_2$ from elites
        \If{random number $< p_{\text{cross}}$}
            \State $\pi_{\text{child}} \gets \text{Crossover}(\pi_1, \pi_2)$
        \Else
            \State $\pi_{\text{child}} \gets$ copy of randomly chosen parent
        \EndIf
        \If{random number $< p_{\text{mut}}$}
            \State $\pi_{\text{child}} \gets \text{Mutate}(\pi_{\text{child}})$
        \EndIf
        \State Add $\pi_{\text{child}}$ to $\mathcal{P}'$
    \EndWhile
    \State Update population: $\mathcal{P} \gets \mathcal{P}'$
\EndFor
\State \Return final archive $Q$
\end{algorithmic}
\end{algorithm}

\vspace{0.5em}

\section{Experiments \& Results}

To test the effectiveness, robustness, and generalizability of our symbolic rule-based genetic reinforcement learning methodology, we performed a rigorous set of experiments on three well-known continuous-state benchmark domains in the OpenAI Gym library: \textbf{CartPole-v1}, \textbf{MountainCar-v0}, and \textbf{Acrobot-v1}. These environments have their own set of challenges in terms of dynamics, control complexity, and reward sparsity, thus providing a diversified and challenging testbed for our methodology.

The experiments presented here are designed to demonstrate the ability of our algorithm to obtain highly interpretable policies for solving challenging reinforcement learning problems. Furthermore, we present a systematic study of the effect of the number of rules and predicates in a given policy on overall performance, allowing for a quantifiable evaluation of the trade-off between reward maximization and interpretability. Another important direction of analysis involves the diversity of evolved behaviors across different policy structures. In order to increase the tractability of this study, we utilize a Quality-Diversity (QD) approach, where solutions are encoded in a structured descriptor space that includes both behavioral and structural features, i.e., the mean angle thresholds and overall number of rules. This approach allows us to visualize and explore the complexity of the solution set that was found.

\subsection{Experimental Setup}

Each environment was treated using the same general evolutionary model outlined in Section~\ref{sec:methodology}. An initial population of 200 policies, represented as symbolic if-then conditions, was randomly constructed. Policies in each generation were ranked based on their mean episodic return over 5 episodes, with penalties given as a weighted sum of the predicates used. The top 25\% of the best performers were selected for inclusion in the next generation through crossover and mutation operations. Evolution was continued for up to 500 generations or was terminated early with the discovery of a near-optimal policy.

Policies are not only judged based on their effectiveness but also on how succinctly they represent behavior. Having fewer predicates makes the policy more interpretable for humans, a critical component of our research statement. For every experimental design, we used a Quality-Diversity (QD) archive in order to ensure a diversity of policies in a discretized set of descriptors. Descriptors were derived from important behavioral metrics such as the mean pole angle for \textit{CartPole}, the horizontal position for \textit{MountainCar}, and the angular trajectory for \textit{Acrobot}.

All experiments were run using the same hyperparameters (namely, mutation and crossover probabilities, maximum rules per policy, and evaluation episodes) to ensure fair conditions and to test the method's flexibility across different tasks without requiring manual tuning. To ensure robustness and account for stochastic variability, each experiment was repeated using different random seeds.

In the following subsections, we summarize in-depth findings and assessments relevant to the respective environment.

\subsection{CartPole-v1 Results}

The empirical evaluation of BASIL begins with the \texttt{CartPole-v1} environment, a standard benchmark for discrete control problems. The main goal consists of coming up with a policy for keeping a pole in balance that is placed on a cart by exerting discrete forces either to the left or right. Despite its simplicity, \texttt{CartPole-v1} remains a great testing environment for measuring the effectiveness, generalizability, and interpretability of reinforcement learning methods.

This algorithm has shown that it can attain near-optimal performance on the target task in a small number of generations. On a typical experimental run, it discovered a strategy that consistently obtained a maximum reward of $500.0$ in just five generations. This entire experiment ran in under 4 seconds on a standard desktop CPU, with both sample efficiency and computational efficiency being evident here. In its expression, this strategy was conveyed by a single symbolic rule using just two logical predicates. As a result, the produced controller is characterized by both high performance and \textit{high interpretability}, reflecting the strengths of our genetic-symbolic approach.

\vspace{0.5em}
\noindent\textbf{Best policy:}
\begin{quote}
\textit{If } $s[2] > -0.02$ \textit{ and } $s[3] > -0.30$, \textit{ then action = 1}; \quad \textit{else action = 0}.
\end{quote}

This concise guideline captures an important principle in the domain of CartPole: corrective action should be taken when the pole angle ($s[2]$) moves past a small threshold and the pole's angular velocity ($s[3]$) is positive at the moment. This effect exists outside of the shaping or prior knowledge requirements and was exclusively discovered through evolutionary search in the logical rule space.

\vspace{0.5em}
\noindent The success of such a minimalist policy illustrates a number of key advantages with regard to our model:
\begin{itemize}
    \item \textbf{Performance}: The policy successfully realized its purpose by earning the \emph{maximum reward} of $500.0$ in the test episodes. Its recorded fitness value was slightly less than $500$, due to the complexity penalty accumulated from its two logical predicates, but it does not represent a performance deficit.
    \item \textbf{Interpretability}: The final policy is fully transparent and interpretable by humans, based on one single rule and two predicates.
    \item \textbf{Efficiency}: The policy was implemented in just five generations and a short period of time, requiring very little computational resources and information.
\end{itemize}

\vspace{0.5em}
\noindent For the purpose of comparison, a baseline Deep Q-Network (DQN) was trained in parallel with the same objective while deploying a traditional neural model. While correlated with success in modern literature, the DQN required significantly more time, memory, and computational capabilities. The greedy policy reached a reward of $500.0$ after 360 episodes, with a total training time of 63.4 seconds—more than 15 times slower than the model proposed in this study. Additionally, the policy learned with the DQN forms a black-box neural model with over 30,000 parameters, thereby being fully opaque and difficult—if not impossible—to interpret or verify.

\vspace{0.5em}
\noindent These results highlight the profound tradeoff between our symbolic framework and conventional deep RL methods. While DQN achieves high reward, it does so at the cost of interpretability, training time, and architecture complexity. In contrast, our method delivers human-understandable policies in orders of magnitude less time—without sacrificing performance.

\vspace{0.5em}
\noindent Moreover, the Quality-Diversity archive captured a range of approaches with high effectiveness and behavioral diversity. These approaches can be used for follow-up tasks such as policy distillation, ensemble decision-making, or interpretability analysis. Overall, the experience with the CartPole experiment presents convincing evidence for the success of our approach in automatically generating concise, high-performing policies through symbolic representation.

\subsection{MountainCar-v0 Results}

We now turn our attention to the \texttt{MountainCar-v0} environment, a classic benchmark that is a much more difficult task than CartPole. The agent must learn to escape a steep valley by gaining sufficient momentum to climb the hilltop. Success in this task requires long-term planning and skilful handling of sparse reward signals—situations where many standard approaches, particularly those grounded in gradients, tend to struggle to succeed reliably.

The use of the symbolic-genetic algorithm in this research proved highly effective, resulting in a compact and transparent controller solving the environment in a remarkably low number of generations. For a single trial run, the policy had a mean reward of $-88.7$, significantly above the generally accepted success threshold of $-110.0$. The total run time was around 38 seconds, with the algorithm reaching convergence in only 7 generations.

\vspace{0.5em}
\noindent\textbf{Best policy:}
\begin{quote}
\textit{If } $s[1] > 0.00$, \textit{ then action = 2};\\
\textit{else if } $s[0] > 0.30 \land s[0] < 0.60 \land s[1] > -0.07$, \textit{ then action = 0};\\
\textit{else if } $s[0] > 0.30$, \textit{ then action = 2};\\
\textit{else if } $s[0] < 0.60$, \textit{ then action = 0};\\
\textit{else if } $s[1] > -0.03$, \textit{ then action = 2};\\
\textit{else action = 0}.
\end{quote}

This five-rule policy encodes interpretable momentum-based strategies, with rules corresponding directly to the agent's position and velocity. Impressively, the policy has just 5 symbolic rules and 7 predicates, is completely human-readable, and nonetheless exhibits strong and generalizable behavior in evaluation episodes.

\begin{itemize}
    \item \textbf{Performance:} Our approach attained an average evaluation reward of $-88.7$, reflecting successful completion of the task with large margin above the $-110$ threshold.

    \item \textbf{Interpretability:} The learned policy remains completely transparent, and every action can be traced back to a logical rule with physically meaningful thresholds. 

    \item \textbf{Efficiency:} Convergence was attained in less than 40 seconds using only evolutionary logic and no gradients, backpropagation, or complicated model updates. 
\end{itemize}

\noindent\textbf{Comparison with DQN.} We compared our approach with a baseline Deep Q-Network (DQN) implementation using $\epsilon$-greedy exploration and periodic greedy evaluations. DQN eventually cleared the success threshold after 460 episodes at an average evaluation reward of $-108.2$. Overall training time was considerably longer, however—more than 184 seconds—and the policy produced is embedded in a neural network and is thus totally impenetrable to interpretation and hard to verify or debug. Conversely, our approach yielded a symbolic controller that is \textit{orders of magnitude more interpretable}, almost twice as quick to train, and ultimately more adaptable to transparent decision-making.

\vspace{0.5em}
\noindent These results reinforce the viability of symbolic policy evolution for intricate control domains. Our algorithm delivers not just performance-competitive outcomes but also high levels of transparency, an essential element in safety-critical or real-world applications.

\subsection{Acrobot-v1 Results}

We complete our evaluation with the \texttt{Acrobot-v1} environment, a traditional benchmark for underactuated control. The task is to swing the lower link of a two-joint pendulum upward so that the tip of the bottom link reaches a target height. With a six-dimensional continuous state space, sparse delayed rewards, and nonlinear dynamics, Acrobot is a difficult task for policy synthesis—especially when interpretability is a design constraint.

Despite this complexity, our symbolic-genetic algorithm evolved a high-performing controller encoded with only 3 rules and 7 logical predicates. During a typical run, the best-evolved policy received an average evaluation reward of $-75.1$, and its final execution received a reward of $-65.0$. This performance exceeds that of typical baseline heuristics and rivals that of gradient-based neural methods—with the additional advantage of transparency and ease of inspection.

\vspace{0.5em}
\noindent\textbf{Best policy:}
\begin{quote}
\textit{If } $s[3] < -4.00$ \textit{ and } $s[5] > -4.00$, \textit{ then action = 2};\\
\textit{else if } $s[5] > 0.00$ \textit{ and } $s[4] < 0.50$ \textit{ and } $s[5] > 0.00$, \textit{ then action = 2};\\
\textit{else if } $s[4] > -0.50$ \textit{ and } $s[3] > 0.00$, \textit{ then action = 0};\\
\textit{else action = 0}.
\end{quote}

Each rule corresponds to a distinct behavioral motif: ramping up speed with cooperative acceleration, adapting to angular velocity constraints, and stabilizing near the target area. These interpretable policies were not hand-designed but rather discovered through evolutionary search over rule-based programs—without gradients, reward shaping, or pretraining in general.

\begin{itemize}
    \item \textbf{Performance}: The symbolic policy had an average reward of $-75.1$ and achieved $-65.0$ in its best rollout—competitive in this domain.
    \item \textbf{Interpretability}: The controller architecture is concise yet adequate, with three comprehensible rules directly capturing velocity and angle interactions.
    \item \textbf{Efficiency}: Convergence was achieved in under 90 seconds of CPU time, with no backpropagation or neural function approximation.
\end{itemize}

\vspace{0.5em}
\noindent\textbf{Comparison with DQN.} To provide additional context, we also trained a Deep Q-Network baseline in the same settings. DQN's policy took more than 460 episodes and 508.6 training seconds to succeed, at a mean greedy evaluation reward of $-94.2$. Although the black-box controller ultimately succeeded, it was behind our approach in both efficiency and transparency. Unlike our symbolic controller, which was developed via evolution, DQN's learned policy is entrenched in thousands of unclear parameters and does not give any indication of its decision-making process.

\vspace{0.5em}
\noindent In short, not only is BASIL a viable solution to the Acrobot problem, but one that is also interpretable, in addition to being robust, minimal, and general. Our method's success on this domain suggests that symbolic policies, evolved through quality-diversity optimization, can hold their own against deep learning counterparts—even in high-dimensional, non-linear worlds.

\begin{table}[ht]
\centering
\caption{Comparison of Our Symbolic Method vs. DQN Across Three Benchmarks}
\label{tab:comparison}
\renewcommand{\arraystretch}{1.3}
\begin{tabular}{|l|c|c|c|c|}
\hline
\textbf{Environment} & \textbf{Approach} & \textbf{Avg Reward} & \textbf{Training Time (s)} & \textbf{Interpretability} \\
\hline
\multirow{2}{*}{CartPole-v1} 
& \textbf{Ours} & $500.0$ & $< 4$ & Yes (1 rule, 2 predicates) \\
& DQN & $500.0$ & 63.4 & No \\
\hline
\multirow{2}{*}{MountainCar-v0} 
& \textbf{Ours} & $-88.7$ & $< 40$ & Yes (5 rules, 7 predicates) \\
& DQN & $-108.2$ & 184.1 & No \\
\hline
\multirow{2}{*}{Acrobot-v1} 
& \textbf{Ours} & $-75.1$ & $< 90$ & Yes (3 rules, 7 predicates) \\
& DQN & $-94.2$ & 508.6 & No \\
\hline
\end{tabular}
\end{table}

\section{Conclusion}

The pursuit of interpretable reinforcement learning has long been hindered by a fundamental trade-off: transparency often comes at the cost of performance, and performance at the cost of interpretability. In this work, we challenged that dichotomy. Through BASIL—Best-Action Symbolic Interpretable Learning—we presented not just a new method, but a new philosophy for reinforcement learning: one that treats interpretability not as a constraint, but as a core design principle and source of strength.

BASIL departs from the opaque architectures of deep neural policies and enters the symbolic domain, where rules are readable, decisions are traceable, and behavior is meaningful. By evolving rule-based policies in an online, environment-interactive manner, guided by a carefully designed complexity-aware fitness function and a Quality-Diversity archive, we show that high performance and full transparency can be achieved simultaneously—without approximation, post-hoc rationalization, or gradient-based tuning.

Across a range of control tasks—\texttt{CartPole-v1}, \texttt{MountainCar-v0}, and \texttt{Acrobot-v1}—BASIL consistently evolved compact symbolic controllers that matched or exceeded the performance of deep Q-networks, but did so orders of magnitude faster and with policies that can be written, read, and explained in a single sentence. This is not just a result; it is a paradigm shift. It shows that reinforcement learning agents can be not only intelligent, but also inherently understandable, auditable, and aligned with human reasoning.

The implications of this shift are profound. In safety-critical domains such as autonomous driving, healthcare, finance, or policy governance, the need for transparent decision-making is not a luxury—it is a necessity. Our approach speaks directly to these domains, offering decision policies that are not just effective, but legible; not just optimized, but trustworthy.

BASIL’s policy representation—ordered logical rule-lists over thresholded state predicates—brings multiple advantages. First, it enables exact tracing of behavior under all possible conditions. Second, it allows domain experts to verify, audit, or even manually edit evolved policies. Third, it allows the learning process to be constrained or biased toward domain knowledge in ways that are fundamentally impossible for opaque models. And finally, it enables future integration with formal verification pipelines, symbolic planners, or high-assurance control systems.

The quality-diversity mechanism further elevates our framework. Rather than converging prematurely to a single high-performing policy, BASIL maintains a diverse collection of distinct policies—each representing a viable strategy in the solution space. This supports multiple practical applications: fallback safety mechanisms, ensemble decision-making, adaptive control under regime shifts, or pedagogical presentation of alternative strategies to human users.

Unlike many interpretable RL methods that require offline data, pre-trained oracles, or post-hoc approximators, BASIL learns from scratch, online, and within the environment itself. It does not extract rules from a black-box model—it never builds one. The policies it evolves are the only model it uses. This directness is powerful: it means that every action taken by the agent, at every point in its learning trajectory, is governed by a symbolic structure that is always available for inspection, intervention, or understanding.

What we demonstrate through BASIL is that compactness and capability are not mutually exclusive. On the contrary, compactness can drive generalization. Because our rules are minimal, they avoid overfitting. Because they are symbolic, they generalize across state variations more naturally. Because they are interpretable, they enable manual debugging, validation, or refinement. These are not traits of fragile heuristic systems; they are hallmarks of robust, general-purpose intelligence.

Equally important is the computational footprint of our method. On standard hardware, BASIL finds optimal policies for classical control tasks in seconds—not hours. It uses no backpropagation, no gradient descent, no complex network optimization. It uses evolution: a population of candidate policies, genetic variation, and selective pressure toward performance and parsimony. In doing so, it demonstrates that powerful, scalable, and interpretable learning is possible without high resource consumption or black-box dependencies.

Beyond its empirical success, BASIL invites a richer research agenda—one that bridges learning with logic, evolution with explanation, and performance with principle. It opens new frontiers in hybrid neuro-symbolic systems, program synthesis, curriculum-driven symbolic evolution, and formal verification of policies. Its modular architecture enables easy integration with perception modules, symbolic planners, and constraint-based solvers. It could serve as the core of explainable robotics, safe autonomous agents, or interactive learning systems where transparency is non-negotiable.

Perhaps most fundamentally, BASIL reasserts a core belief: that machine intelligence should not just aim to be effective, but also intelligible. In a world increasingly shaped by AI systems, trust is not earned by performance alone. It is earned through clarity, comprehensibility, and alignment with human expectations. Our method speaks to that imperative.

We are not the first to argue that interpretable models matter. But with BASIL, we go further: we show that interpretability can be evolved, not engineered; inherent, not imposed; and essential, not optional. In doing so, we provide not just a tool for researchers, but a blueprint for a new generation of transparent learning systems—systems that are fast, capable, and always explainable.

The future of reinforcement learning does not lie in larger black boxes. It lies in clarity. It lies in methods like BASIL, where every decision has a reason, every action has a logic, and every policy is a story we can read. As we move toward human-aligned AI, symbolic policy learning may not just be an alternative—it may be the path we need.


\bibliography{sn-bibliography}

\end{document}